\documentclass[3p,review,authoryear]{elsarticle/elsarticle} 
\pdfoutput=1

\usepackage{amssymb,amsmath,graphicx}
\usepackage{enumerate}
\usepackage{multirow}
\usepackage{tipa}
\usepackage{tabularx}
\usepackage{tikz}
\usepackage{tikzscale}
\usetikzlibrary{patterns}
\usepackage{dblfloatfix}
\usepackage{attachfile2}
\usepackage{tabu}
\usepackage{multicol}
\usepackage{lipsum}
\usepackage{color}
\usepackage{textcomp}
\usepackage{caption}
\usepackage{subcaption}
\usepackage[binary-units]{siunitx}

    \makeatletter
    \def\ps@pprintTitle{%
      \let\@oddhead\@empty
      \let\@evenhead\@empty
      \let\@oddfoot\@empty
      \let\@evenfoot\@oddfoot
    }
    \makeatother

\newcommand{\ph}{\textipa}

\newcolumntype{"}{@{\hskip\tabcolsep\vrule width 1pt\hskip\tabcolsep}}

\definecolor{ib}{rgb}{0, 0, 0}
\definecolor{ir}{rgb}{0, 0, 0}
\definecolor{mc}{rgb}{0, 0, 0}
\definecolor{mr}{rgb}{0, 0, 0}

\newcommand{\rv}{\color{black}}

\begin{document}

\begin{frontmatter}

\title{On Structured Sparsity of Phonological Posteriors\\ for Linguistic Parsing} 

\author[label1]{Milos Cernak\corref{cor1}}
\ead{milos.cernak@idiap.ch}
\author[label1]{Afsaneh Asaei\corref{cor1}}
\ead{afsaneh.asaei@idiap.ch}
\author[label1,label2]{Herv\'e Bourlard}
\ead{herve.bourlard@idiap.ch}

\address[label1]{Idiap Research Institute, Martigny, Switzerland}
\address[label2]{\'Ecole Polytechnique F\'ed\'erale de Lausanne (EPFL), Switzerland}
\cortext[cor1]{Corresponding authors; both authors contributed equally to this manuscript.}

\begin{abstract}

The speech signal conveys information on different time scales from
short (20--40 ms) time scale or segmental, associated to phonological
and phonetic information to long (150--250 ms) time scale or supra
segmental, associated to syllabic and prosodic information. Linguistic
and neurocognitive studies recognize the \emph{phonological} classes
at segmental level as the essential and invariant representations used
in speech temporal organization.

In the context of speech processing, a deep neural network (DNN) is an
effective computational method to infer the probability of individual
phonological classes from a short segment of speech signal. A vector
of all phonological class probabilities is referred to as
\emph{phonological posterior}. There are only very few classes
comprising a short term speech signal; hence, the phonological
posterior is a sparse vector. Although the phonological posteriors are
estimated at segmental level, we claim that they convey
supra-segmental information. Specifically, we demonstrate that
phonological posteriors are indicative of syllabic and prosodic
events.

Building on findings from converging linguistic evidence on the
gestural model of Articulatory Phonology as well as the neural basis
of speech perception, we hypothesize that phonological posteriors
convey properties of linguistic classes at multiple time scales, and
this information is embedded in their support (index) of active
coefficients. To verify this hypothesis, we obtain a binary
representation of phonological posteriors at the segmental level which
is referred to as first-order sparsity structure; the high-order
structures are obtained by the concatenation of first-order binary
vectors. It is then confirmed that the classification of
supra-segmental linguistic events, the problem known as
\emph{linguistic parsing}, can be achieved with high accuracy using a
simple binary pattern matching of first-order or high-order
structures.

\end{abstract}

\end{frontmatter}

\section{Introduction}

A theory of Articulatory Phonology \citep{Browman86} suggests that an
utterance is described by {\rv temporally overlapped distinctive}
constriction actions of the vocal tract organs, known as 
gestures. Gestures are changes in the vocal tract, such as opening and
closing, widening and narrowing, and they are phonetic in nature
\citep{Fowler15}. Gestures compose units of information 
and can be used to distinguish words in all languages. Recent work
on Articulatory Phonology \citep{Goldstein03} further suggests an
existence of coupling/synchronisation of gestures that influence the
syllable structure of an utterance.



Phonological classes (e.g., \citep{Jakobson56,chomsky68sound}) emerge
during the phonological encoding process  -- the processes of speech
planning for articulation, namely the preparation of an abstract
phonological code and its transformation into speech motor plans that
guide articulation~\citep{Levelt93}. \cite{Stevens08} reviews evidence
about a universal set of phonological classes that consists of
articulator-bound classes and articulator-free classes ([continuant],
[sonorant], [strident]). We support the hypothesis in this work and
consider phonological classes in our work as essential and invariant
acoustic-phonetic elements used in both linguistics and cognitive
neuroscience studies for speech temporal organization. 


In the present paper, we study inferred phonological posterior
features that consist of phonological class probabilities given a
segment of input speech signal. The class-conditional posterior
probabilities are estimated using a Deep Neural Network
(DNN). \cite{Cernak15icassp} introduce the phonological posterior
features for phonological analysis and synthesis, and we hypothesise
their relation to the linguistic gestural model. \cite{Saltzman89}
describe the constriction dynamics model as a computational system
that incorporates the theory of articulatory phonology. This gestural
model defines gestural scores as the temporal activation of each
gesture in an utterance. Thus, we hypothesise that phonological
posteriors are related to gestural scores and that the trajectories of
phonological posteriors correspond to the distal representation of
articulatory gestures.  {\color{ir}In a broader view, we consider the
trajectories of phonological posteriors as articulatory-bound and
articulatory-free gestures. Since gestures are linguistically
relevant} \citep{Liberman00}, we hypothesize that phonological
posteriors should convey supra-segmental information through their
inter-dependency low-dimensional structures. Hence, by characterizing
the structure of phonological posteriors, it should be possible to
perform a top-down linguistic parsing, i.e., by knowing \textit{a
priori} where linguistic boundaries lie.

Previously in~\citep{Asaei15compr}, we have shown that phonological
posteriors admit sparsity structures underlying short-term segmental
representations where the structures are quantified as sparse binary
vectors. In this work, we explore this idea further and consider
trajectories of phonological posteriors for supra-segmental
structures. We show that unique structures (codes) exist for distinct
linguistic classes and identification of these structures enables us
to perform linguistic parsing. The linguistic parsing is thus achieved
through identification of low dimensional sparsity structures of
phonological posteriors followed by binary pattern matching. This idea
is in line with an assumption that physical and cognitive speech
structures are, in fact, the low and high dimensional descriptions of
a single (complex)
system\footnote{http://www.haskins.yale.edu/research/gestural.html}.

Our contribution to advance the study of phonological posteriors is
two-fold: First, we review converging evidence from linguistic and
neural basis of speech {\color{ib}perception}, that support the
hypothesis about phonological posteriors conveying properties of
linguistic classes at multiple time scales. Second, we propose
linguistic parsing based on structured sparsity as low dimensional
characterization of phonological posteriors.


The rest of the paper is organized as
follows. Section~\ref{sec:posteriors} provides review of the definition
and relation of phonological posteriors to the linguistic gestural
model and subsequently to cognitive neuroscience,
Section~\ref{sec:linguistic} introduces linguistic parsing, and
Section~\ref{sec:exp} presents the details of experimental analysis. Finally,
Section~\ref{sec:concl} concludes the paper and discusses the results in a
broader context.

\section{Phonological Class-conditional Posteriors}
\label{sec:posteriors}
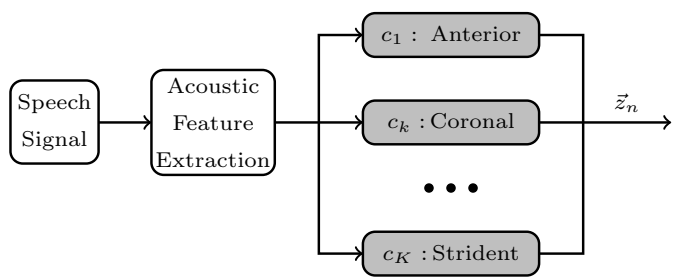
\begin{figure}[!h]
\centering
\resizebox{3.5in}{!}{%
\begin{tikzpicture}[font=\scriptsize]
  \draw[rounded corners, thick] (0,1.28) rectangle (1,2.23);
  \node[align=center] at (0.5,1.75) {Speech\\Signal};

  \draw[rounded corners, thick] (1.60,1.15) rectangle (3,2.35);
  \node[align=center] at (2.3,1.75) {Acoustic\\Feature\\Extraction};
  \draw[thick, ->] (1,1.75) -- (1.60,1.75);
  \draw[thick, ->] (3,1.75) -- (4,1.75);
  \draw[rounded corners, thick, fill=lightgray] (4,2.5) rectangle (6,3);
  \node[align=center] at (5,2.75) {$c_1:\,$ Anterior};

  \draw[thick, <->] (4,2.75) -- (3.5,2.75) -- (3.5,0.25) -- (4,0.25);
  \draw[rounded corners, thick, fill=lightgray] (4,1.5) rectangle (6,2);
  \node[align=center] at (5,1.75) {$c_k:\,$Coronal};
  \draw[thick] (6,2.75) -- (6.5,2.75) -- (6.5,0.25) -- (6,0.25);

  \draw[rounded corners, thick, fill=lightgray] (4,0) rectangle (6,0.5);
  \node[align=center] at (5,0.25) {$c_K:\,$Strident};
  \draw[thick, fill=black] (4.75,1) circle[radius=0.04];
  \draw[thick, fill=black] (5,1) circle[radius=0.04];
  \draw[thick, fill=black] (5.25,1) circle[radius=0.04];

  \draw[thick, ->] (6,1.75) -- (7.5,1.75);
  \node[above] at (7,1.75)  {$\vec{z}_n$};
\end{tikzpicture}
}%
\caption{\emph{The process of phonological analysis. Each segment of
    speech signal is represented by phonological posterior
    probabilities $\vec{z}_n$ that consist of $K$ class-conditional
    posterior probabilities. For each phonological class, a DNN is
    trained to estimate its posterior probability given the input
    acoustic features.}}
\label{fig:phonovocTRN}
\end{figure}

Figure~\ref{fig:phonovocTRN} illustrates the process of the
phonological analysis \citep{Yu2012,Cernak15icassp}. The phonological
posterior features are extracted starting with converting a segment of
speech samples into a sequence of acoustic features
$X=\{\vec{x}_1,\ldots,\vec{x}_n,\ldots,\vec{x}_N\}$ where $N$ denotes
the number of segments in the utterance. Conventional cepstral
coefficients can be used as acoustic features. Then, a bank of
phonological class analysers realised via neural network classifiers
converts the acoustic feature observation  sequence $X$ into a
sequence of phonological posterior probabilities
$Z=\{\vec{z}_1,\ldots,\vec{z}_n,\ldots,\vec{z}_N\}$; a posterior
probability
$\vec{z}_n=[p(c_1|x_n),\hdots,p(c_k|x_n),\hdots,p(c_K|x_n)]^\top$
consists of $K$ phonological class-conditional posterior probabilities
where $c_k$ denotes the phonological class and $.^\top$ stands for the
transpose operator.

\begin{figure}
\centering
  \includegraphics[width=.97\linewidth,height=.40\linewidth]{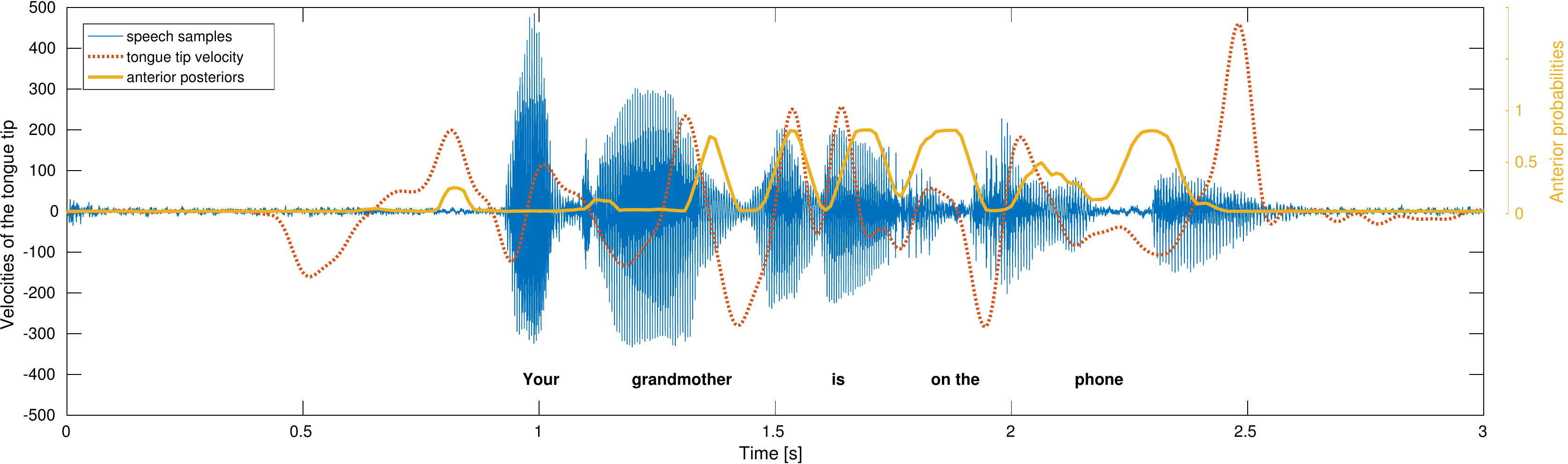} 
  \caption{{\it Anterior phonological posteriors vs. the
      electromagnetic articulography tongue tip measurement. {\rv
        Anterior posteriors were generated using analysis described in
        Section~\ref{sec:analysis}}. The correlation between the 
      articulatory gesture score and the trajectory of its
      corresponding phonological class posterior probability is
      evident.}}
\label{fig:tt}
\end{figure}

The phonological posteriors $Z$ yield a parametric speech representation, and 
we hypothesise that the trajectories of the articulatory-bound
phonological posteriors correspond to the distal representation of the
gestures in the gestural model of speech production (and
perception). For example, Figure~\ref{fig:tt} shows a comparison of
articulatory tongue tip gestures (vertical direction with respect to
the occlusal plane) and the phonological anterior posterior features,
on an electromagnetic articulography (EMA)
recording~\citep{lee05}. The articulatory gesture and 
phonological posteriors trajectory have the same number of maximums,
and their relation is evident. {\rv A more asynchronous relation
  toward the end of utterance is caused by asynchronous relations of
  fast tongue tip movements causing an obstruction in the mouth and
  generated acoustics.}

The hypothesis of correspondence of the phonological posterior
features to the gestural trajectories is also motivated by the analogy
to the constriction dynamics model \citep{Saltzman89} that takes
gestural scores as input and generates articulator trajectories and
acoustic output. Alternatively to this constriction dynamics model, we
generate acoustic output using a phonological synthesis DNN described
in~\cite{Cernak15icassp}.

In the following sections, we outline converging evidence from
linguistics as well as the neural basis of speech
{\color{ib}perception}, that support the hypothesis  about
phonological posteriors conveying properties of linguistic classes at
multiple time scales.


\subsection{Linguistic Evidence}
\label{sec:lingevidence}


Linguistics defines two traditional components of speech structures:
\begin{enumerate}
\item Cognitive structure consisting of system primitives, that is,
  the units of representation for cognitively relevant objects such
  as phonemes or syllables. The system primitives are represented by
  \emph{canonical} phonological features (classes) that emerge during
  the phonological encoding process~\citep{Levelt93}.
\item Physical structure generated by a set of permissible operations
  over cognitive system primitives that yield the observed (surface)
  patterns. The physical structure is represented by \emph{surface}
  phonological features, continuous variables that may be partially
  estimated from the speech signal by inverse filtering. Phonological
  posteriors can also be classified as surface phonological features.
\end{enumerate}


The canonical (discrete) phonological features have been used over the
last 60 years to describe cognitive structures of speech 
sounds. \cite{Miller55} have experimentally shown that consonant {\rv
confusions were perceived similarly for the isolated phonemes and the
phonemes within the perceptual groups they form (for example /\ph{t}/,
/\ph{p}/ and /\ph{k}/ form a perceptual group, characterised by
binary features of voiceless stops).}
Canonical features are extensively studied in
phonology. In the tradition of~\cite{Jakobson56}
and~\cite{chomsky68sound}, phonemes are assumed to consist of feature
bundles -- the Sound Pattern of English (SPE). Later advanced
phonological systems were proposed,  such as multi-valued phonological
features of \cite{Ladefoged14}, and monovalent Government Phonology
features of \cite{Harris95} that describe sounds by fusing and
splitting of primes.


The surface code includes co-articulated canonical code, with further
intrinsic (speaker-based) and extrinsic (channel-based) speech
variabilities that contribute to the opacity of the function operating
between the two codes. The surface features may contain additional
gestures dependent on the prosodic context, such as position within a
syllable, word, and sentence. Other changes in surface phonological
features at different time granularities are due to phonotactic
constraints. For example, glides are always syllable-initial, and
consonants that follow a non-tense vowel are always in the coda of the
syllable~\citep{Stevens08}.

Relation of the canonical and surface code can be investigated by a
linguistic theory of Articulatory
Phonology~\citep{Browman86,Browman89,Browman92} that introduced
articulatory gestures as {\rv a} basis for human speech
production. Although it is generally claimed that gestures convey
segmental-level information (for example, \cite{Fowler15} say that
gestures are phonetic in nature), recent developments suggest that the
timing of articulatory gestures encodes syllabic (and thus linguistic)
information as well~\citep{Browman88,Nam09}. \cite{Liberman00}
provides theoretic claims about {\rv linguisticlly} relevant
articulatory gestures, and \cite{Saltzman89} implement a syllable
structure-based gesture coupling model. Thus, {\rv from prior evidence that articulatory gestures convey linguistic properties, and} the
hypothesis of correspondence of the phonological posteriors to the
gestural trajectories, we claim that phonological posteriors may
convey linguistic information as well.
\newline


\subsection{Cognitive Neuroscience Evidence}



Modern cognitive neuroscience studies use phonological classes as
essential and invariant acoustic-phonetic primitives for speech
temporal organization \citep{Poeppel14}. Neurological data from the
brain activity during speech planning, production or perception are
increasingly used to inform such cognitive models of speech and
language. 

The auditory pre-processing is done in the cochlea,
and then split into two parallel pathways
leading from the auditory system \citep{Wernicke1874}. For example,
the dual-stream model of the functional anatomy of language 
\citep{Hickok07} consists of a ventral stream: sound to meaning
function using phonological classes, phonological-level processing
at superior temporal sulcus bilaterally, and a dorsal stream: sound
to action, a direct link between sensory and motor representations of
speech based again on the phonological classes. The
former stream supports the speech perception, and the latter stream
reflects the observed disruptive effects of altered auditory feedback
on speech production. \cite{Phillips00,Mesgarani2014} present evidence
of discrete phonological classes available in the human auditory
cortex.


Recent evidence from psychoacoustics and neuroimaging studies indicate
that auditory cortex segregates information emerging from the cochlea
on at least three discrete time-scales processed in the auditory
cortical hierarchy: (1) ``stress'' $\delta$ frequency (1--3 Hz), (2)
``syllabic'' $\theta$ frequency (4--8 Hz) and (3) ``phonetic'' low
$\gamma$ frequency (25--35 Hz) \citep{Giraud12}. \cite{Leong14jasa}
show that phase relations between the phonetic and syllabic amplitude
modulations, known as hierarchical phase locking and nesting or
synchronization across different temporal
granularity~\citep{Lakatos05}, is a good indication of the syllable
stress. Intelligible speech representation with stress and accent
information can be constructed by asynchronous fusion of phonetic and
syllabic information~\citep{Cernak15ieee}.


In addition, not only phase locking across different temporal
granularity has linguistic interpretation. \cite{Bouchard13} claim
that functional organisation of ventral sensorimotor cortex supports
the gestural model developed in Articulatory Phonology. Analysis of
spatial patterns of activity showed a hierarchy of network states that
organizes phonemes by articulatory-bound phonological
features. \cite{Leonard2015} further show how listeners use
phonotactic knowledge (phoneme sequence statistics) to process spoken
input and to link low-level acoustic representations (the
coarticulatory dynamics of the sounds through the encoding of
combination of phonological features) with linguistic information
about word identity and meaning.
This is converging evidence on the relation of the linguistic
gestural model and speech and language cognitive neuroscience models
with the phonological class-conditional posteriors used in our work.

\section{Sparse Phonological Structures for Linguistic
  Parsing}\label{sec:linguistic}
\begin{figure}[t]
\centering
  \includegraphics[width=.7\linewidth]{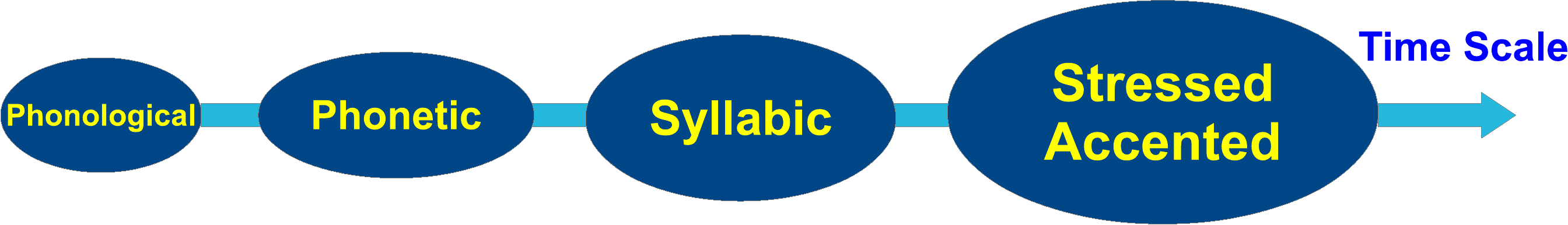} %
  \caption{{\it Different time granularity of speech processing. The
      phonological and phonetic classes are segmental attributes
      whereas the syllable type, stress and accent are linguistic
      events recognized at supra-segmental level. Inferring the
      supra-segmental attributes from sub-phonetic features is the
      task of linguistic parsing~\citep{Poeppel03}. This paper
      demonstrates that phonological posteriors are indicative of
      supra-segmental attributes such as syllables, stress and
      accent.}}
  \label{fig:parsing}
\end{figure}

{\color{ib}Building on linguistic and cognitive findings}, the
phonological representation of speech lies at the center of human
speech processing. Speech analysis is performed at different time
granularities broadly categorized as segmental and supra-segmental
levels. The phonological classes define the sub-phonetic and phonetic
attributes recognized at the segmental level whereas the syllables,
lexical stress and prosodic accent are the basic supra-segmental
events - c.f. Figure~\ref{fig:parsing}. The phonological
representations are often studied at segmental level and their
supra-segmental properties are not investigated. This
supra-segmental characterization of phonological posteriors will be explored in this work.


\subsection{Structured Sparsity of Phonological Posteriors}
Phonological posteriors are indicators of the physiological posture of
human articulation machinery. Due to the physical constraints, only
few combinations can be realized in our vocalization. This physical
limitation leads to a small number of unique patterns exhibited over
the entire speech corpora~\citep{Asaei15compr}. We refer to this
structure as \emph{first-order structure} which is exhibited at the
segmental level.
 
Moreover, the dynamics of the structured sparsity patterns is slower
than the short segments and it is indicative of supra-segmental
information, leading to a higher order structure underlying a sequence
(trajectory) of phonological posteriors. This structure is exhibited
at supra-segmental level by analyzing a long duration of phonological
posteriors, and it is associated to the syllabic information or more
abstract linguistic attributes. We refer to this structure as
\emph{high-order structure}.

We hypothesize that the first-order and high-order structures
underlying phonological posteriors can be exploited as indicators of
supra-segmental linguistic events. To test this hypothesis, we
identify all structures exhibited in different linguistic classes. The
set of class-specific structures is referred to as the codebook.

\subsection{Codebook of Linguistic Structures}\label{sec:codebook}
The goal of codebook construction is to collect all the structures
associated to a particular linguistic event. To that end, we consider
\emph{binary} phonological posteriors where the probabilities above
0.5 are normalized to 1 and the probabilities less than 0.5 are forced
to zero. This rounding procedure enables us to identify the active
phonological components as indicators of linguistic events. It also
alleviates the speaker and environmental  variability encoded in the
continuous probabilities. An immediate extension to this approach is
multi-valued quantization of phonological posteriors as opposed to
1-bit quantization. We consider this extension for our future studies
and focus on binary phonological indicators to obtain linguistic
structures. 

Different codebooks are constructed for different classes. Namely, one
codebook encapsulates all the binary structures of the consonants
whereas another codebook has all the binary structures of the
vowels. These two codebooks will be used for binary pattern matching
to classify consonants versus vowels as will be explained in the next
Section~\ref{sec:match}. Likewise, one codebook encapsulates all the
binary structures of stressed syllables whereas another codebook has
all the binary structures of unstressed syllables, and these two
codebooks are used for stress detection; a similar procedure holds for
accent detection.

The codebook can be constructed from the first-order structures as
well as the high-order structures. For example, a second-order
codebook is formed from all the binary structures of second-order
phonological posteriors obtained by concatenation of two adjacent
phonological posteriors to form a super vector from the segmental
representations.

The procedure of codebook construction for classification of
linguistic events relies on the assumption that there are unique
structures per class (consonant, stressed or syllable) and the number
of permissible patterns is small {\rv(see~\ref{sec:class-specific} for empirical results)}. Hence, classification of any
phonological posterior can be performed by finding the closest match
to its binary structure from the codebooks characterizing different
linguistic classes.

\subsection{Pattern Matching for Linguistic Parsing}\label{sec:match}
Figure~\ref{fig:parsing} illustrates different time granularity
identified for processing of speech. Inferring the supra-segmental
properties such as syllable type or accented / stressed pronunciation
is known as linguistic parsing~\citep{Poeppel03}. Parsing can be
performed in a top-down procedure, driven by a-priori known segment
boundaries.


Having the codebooks of structures underlying phonological posteriors,
linguistic parsing amounts to binary pattern matching. The similarity
metric plays a critical role in classification accuracy. Hence, we
investigate several metrics found effective in different binary
classification settings. The definition of binary similarity measures
are expressed by \emph{operational taxonomic
  units}~\citep{Taxonomy}. Consider two binary vectors $i,\, j$: $a$
denotes the number of elements where the values of both $i,\,j$ are 1,
meaning \emph{``positive match''}; $b$ denotes the number of elements
where the values of $i,\;j$ is $(0,1)$, meaning \emph{``$i$ absence
  mismatch''}; $c$ denotes the number of elements where the values of
$i,\;j$ is $(1,0)$, meaning \emph{``$j$ absence mismatch''}; $d$
denotes the number of elements where the values of both $i,\,j$ are 0,
meaning \emph{``negative match''}.
The definition of binary similarity measures used for our evaluation
of linguistic parsing is as follows~\citep{Choi10asurvey}:

\begin{align}
 &S_{\text{JACCARD}}=\frac{a}{a+b+c}\\ \label{eq:jac}
 &S_{\text{INNERPRODUCT}}=a+d\\
 &S_{\text{HAMMING}}=b+c\\
 &S_{\text{AMPLE}}=\frac{a(c+d)}{c(a+b)}\\
 &S_{\text{SIMPSON}}=\frac{a}{\text{min}(a+b,a+c)}\\
 &S_{\text{HELLINGER}}=2\sqrt{1-\frac{a}{\sqrt{(a+b)(a+c)}}}
\end{align}
Different metrics are motivated due to different treatment of
positive/negative match and mismatches in indicators of phonological
classes. The most effective similarity measure for linguistic parsing
can imply different cognitive mechanisms governing the human
perception of linguistic attributes.
 
In the top-down approach to linguistic parsing, syllable boundaries
are first estimated from the speech signal. Then, the similarity
between the class-specific codebook members and a phonological
posterior is measured. The class label is determined based on the
maximum similarity. We provide empirical results on linguistic parsing
in the following Section~\ref{sec:exp}.

\section{Experiments}
\label{sec:exp}

\subsection{Experimental setup}
We use an open-source {\rv phonological vocoding
platform~\citep{Cernak2016phonvoc} to obtain
phonological posteriors}. Briefly, the platform is based on cascaded
speech analysis and synthesis that works internally with the
phonological speech representation. In the phonological analysis part,
phonological posteriors are detected directly from the speech signal
by DNNs. {\rv
  Binary~\citep{Yu2012} or multi-valued
  classification~\citep{Stouten06,Rasipuram11} might be used. In the
  latter case, the phonological classes are grouped together
  based on place or manner of articulation. We followed the binary
  classification approach in our work, and thus each DNN determines
  the probability  of a particular phonological class.} To confirm
independence of the proposed methodology on a phonological system, two
different phonological speech representations are considered: the SPE
feature set~\citep{chomsky68sound}, and the extended SPE (eSPE)
feature set~\citep{Cernak15icassp} are used in training of the DNNs
for phonological posterior estimation on English and French data
respectively. The mapping used to map from phonemes to SPE
phonological class is taken from~\cite{Cernak2016a}. The distribution
of the phonological labels is non-uniform, driven by mapping different
numbers of phonemes to the phonological classes.

For French eSPE feature set, we started from pseudo-phonological
feature classification designed for American English~\citep{Yu2012}. We
deleted the glottal and dental classes corresponding to English phonemes
\ph{[h, D, T]}, replaced [+Retroflex] with [+Uvular] to represent a
French rhotic consonant, and replaced the broad classes [+Continuant,
+Tense] with:
\begin{itemize}
  \item \emph{Fortis} and \emph{Lenis}, as an alternative to [+Tense]
    class, to distinguish consonants produced with greater and lesser
    energy, or articulation strength.
  \item \emph{Alveolar} and \emph{Postalveolar}, to distinguish
    between sibilants articulated by anterior portion of the tongue,
  \item \emph{Dorsal}, to group consonants articulated by the central
    and posterior portions of the tongue,
  \item \emph{Central}, to group vowels in the central position of the
    portion of tongue that is involved in the articulation and to the
    tongue's position relative to the palate~\citep{Bauman2011},
  \item \emph{Unround}, to group vowels with an opposite degree of lip
    rounding to the [+Round] class.
\end{itemize}

In the following, we describe the databases and DNN training procedure
to estimate the phonological posterior features. 

\subsubsection{Speech Databases}

{\rv To confirm cross-lingual property of phonological
  posteriors~\citep{Siniscalchi2012}}, we conducted our evaluations on
English and French speech corpora. Table~\ref{tab:data} lists data
used in the experimental setup. 

\begin{table}[th]
  \caption{\label{tab:data} {\it Data used for DNN training to obtain
      phonological posteriors, and evaluation data.}}
\vspace{2mm}
\centerline{
\begin{tabu}{|l|c|c|c|}
\hline 
Purpose & Database & Size (hours) \\
\hline  \hline
Training English data & WSJ & 66 \\
Training French data & Ester & 58\\
\hline
Evaluation English data & Nancy &  1.5 \\
Evaluation French data & SIWIS &  1 \\
\hline
\end{tabu}}
\end{table}

To train the DNNs for phonological posterior estimation on English
data, we use the Wall Street Journal (WSJ0 and WSJ1) continuous speech
recognition corpora \citep{WSJDB}. To train the DNNs for phonological
posterior estimation on French data, we use the Ester
database~\citep{ester06} containing standard French radio broadcast
news in various recording conditions.

Once DNNs are trained, the phonological posterior features are estimated for
the Nancy and SIWIS recordings which is used for the subsequent
cross-database linguistic parsing experiments.

The Nancy database is provided in Blizzard
Challenge\footnote{\url{http://www.cstr.ed.ac.uk/projects/blizzard/2011/lessac_blizzard2011}}. The
speaker is known as ``Nancy'', and she is a US English native female
speaker. The database consists of 16.6 hours of high quality
recordings of natural expressive human speech 
recorded in an anechoic chamber at a \SI{96}{\kHz} sampling rate.
The
audio of the last 1.5 hours of the recordings was selected and
re-sampled to sampling frequency of \SI{16}{\kHz} for our
experiments. The transcription of the audio data comprised of {\rv
  \num{12095} utterances}. The text was processed by a conventional
and freely available text to speech synthesis (TTS)
front-end~\citep{festival}, resulting in segmental (quinphone 
phonetic context) and supra-segmental (full-context) labels. The
full-context labels included binary lexical stress and prosodic
accents. In this work, by the term stress, we refer to the lexical
stress of a word, which is the stress placed on syllables within
words. On the other hand, accent refers to the phrase- or sentence-
level prominence given to a syllable. The syllables conveying phrasal
prominence are called pitch accented syllables. In some cases, a
stressed syllable can be promoted to pitch accented syllable based
just on it’s position in a phrase or on the focus/emphasis the speaker
intends to give to the specific part of the sentence to convey a
specific message~\citep{Gordon2014}. Accents are predicted from the text transcriptions, using features that affect accenting, such as
lexical stress, part-of-speech, and ToBI labels. The labels were force-aligned with the audio recordings.

The SIWIS
database\footnote{\url{https://www.idiap.ch/project/siwis/downloads/siwis-database}}
consists of 26 native French speakers. The labels were obtained using
forced alignment. We generated full-context labels using the French
text analyzer eLite~\citep{Roekhaut2014}. Unlike Nancy speech
recordings, SIWIS data is noisy and recorded in less restricted
acoustic conditions. Evaluations on both English and French
corpora enables us to confirm and compare the applicability of our
linguistic parsing method across languages with different phonological
classes as well as different recording scenarios. 


\subsubsection{DNN Training for Phonological Posterior Estimation}
\label{sec:analysis}
First, we trained a phoneme-based automatic speech recognition system
using mel frequency cepstral coefficients (MFCC) as acoustic
features. The phoneme set comprising of 40 phonemes
(including ``sil'', representing silence) was defined by the CMU
pronunciation dictionary. The three-state, cross-word triphone models
were trained with the {\rv HMM-based speech synthesis system (HTS)}
variant~\citep{Zen:HTS} of the {\rv Hidden Markov  Model Toolkit
  (HTK)} on the 90\% subset of the {\rv WSJ} \textit{si\_tr\_s\_284}
set. The remaining 10\% subset was used for cross-validation. We tied
triphone models with decision tree state clustering based on the
minimum description length (MDL) criterion~\citep{shinoda:mdl}. The
MDL criterion allows an unsupervised determination of the number of
states. We used \num{12685} tied models, and modeled each state with a
GMM consisting of 16 Gaussians. The acoustic models were used to get
boundaries of the phoneme labels.

Then, the labels of phonemes were mapped to the SPE phonological
classes. In total, $K$ DNNs were trained as the phonological analyzers
using the short segment (frame) alignment with two output labels
indicating whether the $k$-th phonological class exists for the
aligned phoneme or not. {\rv In other words, the two DNN outputs
  correspond to the target class v/s the rest. Some classes might seem
  to have unbalanced training data, for example, the two labels for
  the nasal class are associated with the speech samples from just 3
  phonemes /\ph{m}/, /\ph{n}/, and /\ph{N}/, and with the remaining 36 
phonemes. However, this split is necessary to train a discriminative
classifier well.
Each DNN was trained on
  the whole training set.} The number of classifiers $K$ is determined from the
set of phonological classes and it is equal to 15 for the English
data, and 24 for the French data.  
The DNNs have the architecture of $351 \times 1024 \times 1024 \times
1024 \times 2$ neurons,
determined  empirically (we tried different sizes, from 500 to
\num{2000}, and deep, from 3 to 5 hidden layers, and different
context of the input features).
The input vectors are 39 order MFCC features
with the temporal context of 9 successive frames.
The parameters were initialized using deep belief network pre-training
done by single-step contrastive divergence (CD-1) procedure of
\cite{Hinton06} on randomly selected \num{5000} utterances from the
training data. The DNNs with the
softmax output function were then trained using a mini-batch based
stochastic gradient descent algorithm with the cross-entropy cost
function of the KALDI
toolkit~\citep{PoveyASRU2011}. Table~\ref{tab:SPEaccuracies} lists the
detection accuracy for different phonological classes. The {\rv DNN}
outputs provide the phonological posterior probabilities for each
phonological class. Detection accuracy on cross-database (Nancy) test
data is lower, however following the accuracy on training and CV
data. The consonantal and voice DNN performed worse, and we
speculate that this might be caused by difficulty to train good
classifiers for such broad {\rv (hierarchical) phonological
  classes. Recent work of~\cite{Nagamine15} suggests that a DNN learns
  selective phonological classes in different nodes and hidden
  layers. Thus, using features from the hidden layers could improve
  classification accuracy of some specific phonological classes.}


{\small
\begin{table}[h]
  \caption{\label{tab:SPEaccuracies} {\it Classification accuracies (\%)
      of English phonological class detectors on train,
      cross-validation (CV) {\color{mc}and cross-database (CDB) test}
    data. {\rv The performance of the silence class, that is
      non-phonological, is not reported.}}}
\vspace{-4mm} 
\centerline{
\begin{tabu}{|l|c|c|c|[0.8pt]l|c|c|c|}
\hline
Phonological & \multicolumn{3}{c|[0.8pt]}{Accuracy (\%)} & Phonological &
\multicolumn{3}{c|}{Accuracy (\%)} \\
\cline{2-4} \cline{6-8}
Classes & Train & CV & CDB & Classes & Train & CV & CDB\\
\hline \hline
vocalic & 97.3 & 96.5 & 95.0 & round & 98.7 & 98.1 & 79.7 \\
consonantal & 96.3 & 95.0 & 78.0 & tense & 96.6 & 95.3 & 83.0 \\
high & 97.0 & 95.7 & 91.8 & voice & 96.5 & 95.6 & 80.5 \\
back & 96.2 & 94.8 & 96.1 & continuant & 97.3 & 96.3 & 92.8 \\
low & 98.4 & 97.6 & 82.5 & nasal & 98.9 & 98.4 & 81.2 \\
anterior & 96.8 & 95.6 & 85.6 & strident & 98.7 & 98.2 & 97.1\\
coronal & 96.1 & 94.6 & 85.2 & rising & 98.6 & 97.8 & 91.8 \\
\hline
\end{tabu}}
\end{table}}

Similarly, we trained French phonological posterior estimators. 
In this study, we retained a subset of Ester consisting of native
speakers in low noise conditions. The three-state, cross-word triphone
models were trained on {\rv 93\% of the data}. The remaining  7\%
subset was used for cross-validation. The acoustic models were used to
get boundaries of the phoneme labels. We tied triphone models with the
MDL criterion that resulted {\rv in} the \num{11504} tied models,
modelling each state with a GMM consisting of 16 Gaussians.

The phoneme set comprising 38 phonemes (including ``sil'') was defined
by the BDLex~\citep{bdlex} lexicon. The aligned phoneme labels were
mapped to the French eSPE phonological classes.
The DNN architecture is similar to the English data, and it is
initialized by deep belief network
pre-training. Table~\ref{tab:eSPEaccuracies} lists the detection
accuracy for various eSPE classes. {\color{mc} Similarly as for
  English, some class detectors evaluated on cross-database (Siwis)
  test data show lower performance. The detectors for the ``broader''
  classes, such as vowel or voiced, perform worse.}

{\small
\begin{table}[h]
  \caption{\label{tab:eSPEaccuracies} {\it Classification accuracies (\%)
      of the French phonological class detectors on train and
      cross-validation (CV) {\color{mc}and cross-database (CDB) test}
      data.{\rv The performance of the silence class, that is
      non-phonological, is not reported.}}}
\vspace{-4mm}
\centerline{
\begin{tabu}{|l|c|c|c|[0.8pt]l|c|c|c|}
\hline
Phonological & \multicolumn{3}{c|[0.8pt]}{Accuracy (\%)} & Phonological &
\multicolumn{3}{c|}{Accuracy (\%)} \\
\cline{2-4} \cline{6-8}
Classes & Train & CV & CDB & Classes & Train & CV & CDB \\
\hline \hline
Labial & 98.2 & 97.4 & 93.5 & Nasal & 99.0 & 98.8 & 86.1 \\
Dorsal & 97.3 & 96.3 & 91.4 & Stop & 97.6 & 97.0 & 90.1 \\
Coronal & 95.9 & 94.7 & 85.0 & Approximant & 98.2 & 97.6 & 95.2 \\
Alveolar & 98.9 & 98.4 & 94.3 & Anterior & 95.4 & 94.2 & 82.5 \\
Postalveolar & 99.7 & 99.5 & 99.0 & Back & 98.0 & 97.1 & 90.6 \\
High & 97.0 & 95.9 & 90.4 & Lenis & 98.0 & 97.4 & 94.3 \\
Low & 97.4 & 96.5 & 84.2 & Fortis & 97.5 & 96.8 & 89.0 \\
Mid & 96.9 & 96.2 & 89.6 & Round & 97.3 & 96.6 & 89.3 \\
Uvular & 98.7 & 98.1 & 94.9 & Unround & 95.9 & 95.1 & 80.6 \\
Velar & 99.2 & 98.8 & 97.5 & Voiced & 95.4 & 94.3 & 80.0 \\
Vowel & 94.3 & 93.1 & 73.0 & Central & 98.5 & 98.1 & 96.6 \\
Fricative & 97.1 & 96.1 & 88.1 & & & & \\
\hline
\end{tabu}}
\end{table}}


\subsection{Linguistic Parsing}
In this section, we present the evaluation results of our proposed
method of top-down linguistic parsing. We provide empirical results on
sparsity of phonological posteriors and confirm validity of
class-specific codebooks to classify supra-segmental linguistic events
based on binary pattern matching.
\subsubsection{Binary Sparsity of Phonological Posteriors}
\label{sec:binary}
\begin{figure}[h]
\centering
\begin{subfigure}{.5\textwidth}
  \centering
  \includegraphics[width=.97\linewidth]{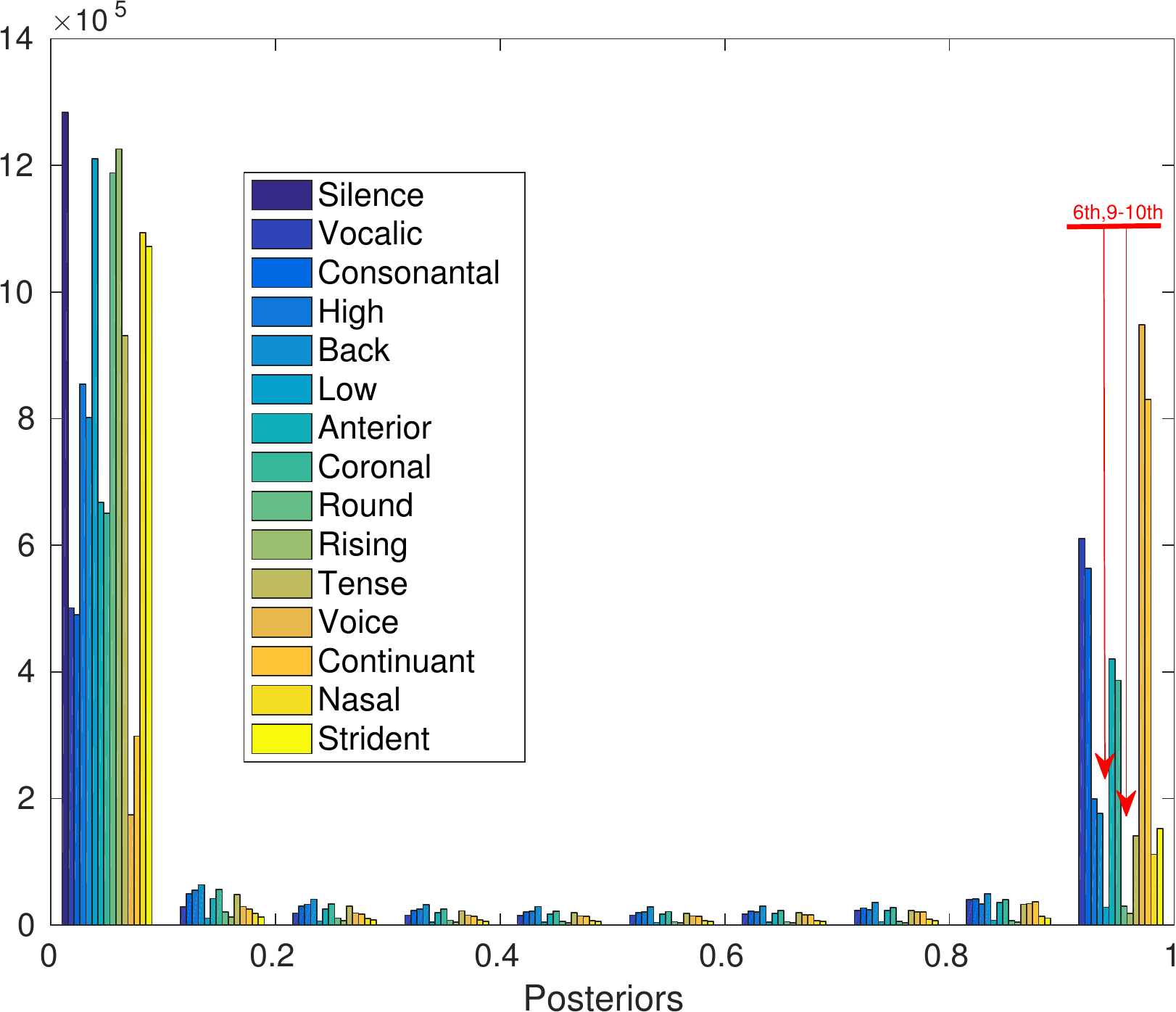}
  \caption{Unstressed Syllables}
  \label{fig:unstressed}
\end{subfigure}%
\begin{subfigure}{.5\textwidth}
  \centering
  \includegraphics[width=.97\linewidth]{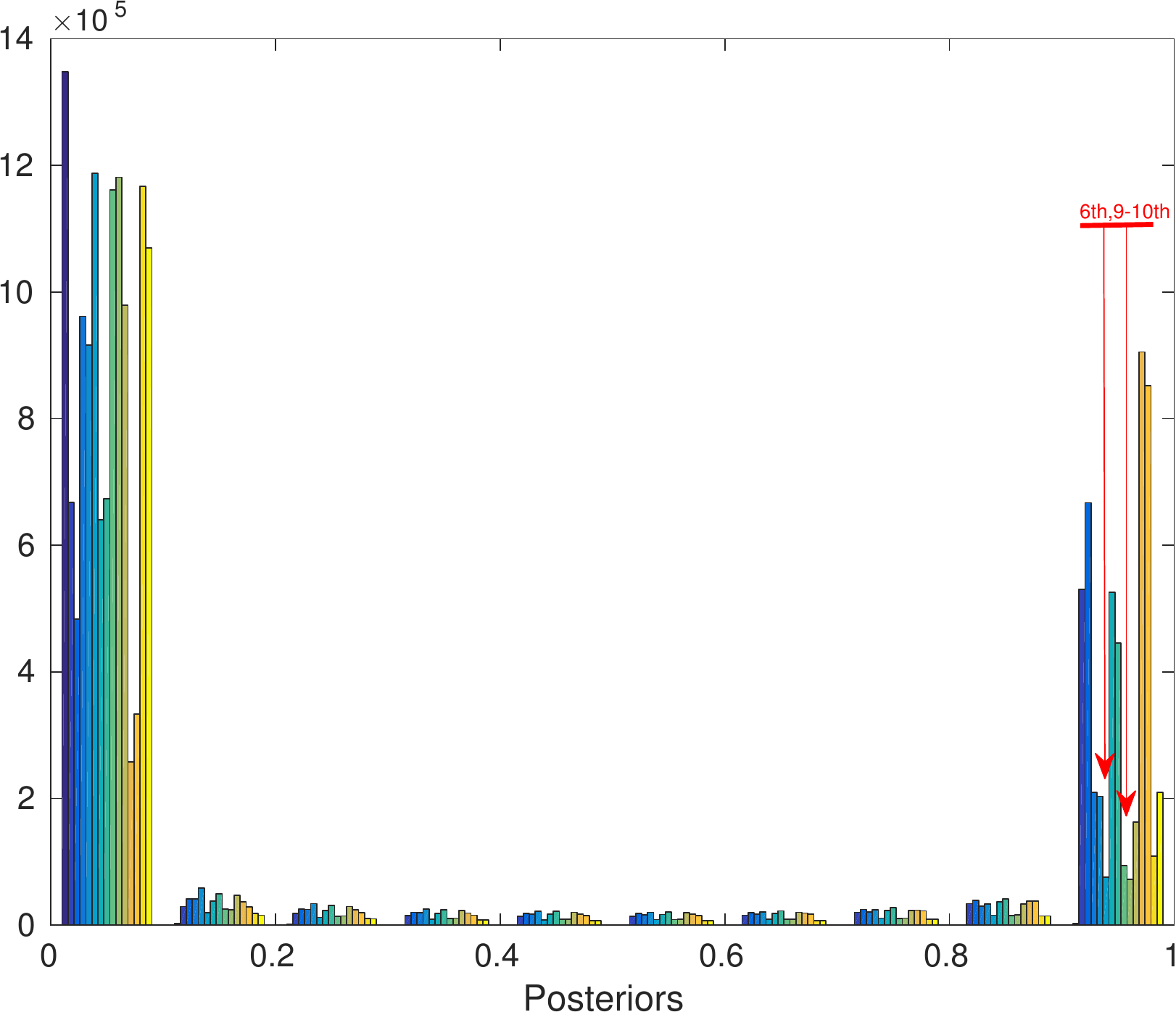}
  \caption{Stressed Syllables}
  \label{fig:stressed}
\end{subfigure}
\caption{{\it Binary sparsity of {\rv English} continuous phonological
    posteriors $Z$. We can observe that at least the [low] (6th),
    [round] (9th) and [rising] (10th) classes are significantly more
    present in stressed binary-ones than in unstressed syllables.}}
\label{fig:binary}
\end{figure}

Figure \ref{fig:binary} illustrates a histogram of phonological
posteriors distribution. We can see that the distribution exhibits the
binary nature of phonological posterior being valued in the range of
$[0-1]$, and mostly concentrated very close to either 1 or 0. This
binary pattern is visible for both stressed and unstressed syllables
as demonstrated in the right and left plots, respectively.

The 1-bit discretization, achieved by rounding of posteriors results
in a very small number of unique phonological binary structures, 0.1\%
of all possible structures. This implies that the binary patterns may
encode particular shapes of the vocal tract. Since a limited number of
these shapes can be created for human speech, the number of unique
patterns is very small.

This property encouraged us also to use this binary approximation in
low bit-rate speech coding~\citep{Cernak15icassp,Asaei15compr}; these
studies confirmed that binary approximation has only a negligible
impact on perceptual speech quality.

Furthermore, comparing Figures~\ref{fig:unstressed} and
\ref{fig:stressed}, we can observe that at least the [low] (6th),
[round] (9th) and [rising] (10th) classes are significantly more
present in stressed binary-ones than in unstressed syllables. This
observation indicates that stressed syllables are more prominent in
prosodic typology~(e.g., \citep{Jun05}). We use
the [rising] feature to differentiate diphthongs from monophthongs,
that is also more prominent in stressed syllables.

%

\subsubsection{Class-specific Linguistic Structures}\label{sec:class-specific}
The objective of this section is to confirm the hypothesis that phonological posteriors admit class-specific structures which can be used for identification of supra-segmental linguistic events. 

Following the procedure of codebook construction elaborated in
Section~\ref{sec:codebook}, we obtain six different codebooks to
address the following parsing scenarios:
\begin{itemize}\itemsep0em
 \item Consonant vs. vowel (C-V) detection.
 \item Stress vs. unstressed detection.
 \item Accented vs. unaccented detection. 
\end{itemize}

\begin{figure}
\centering
  \includegraphics[width=1\linewidth]{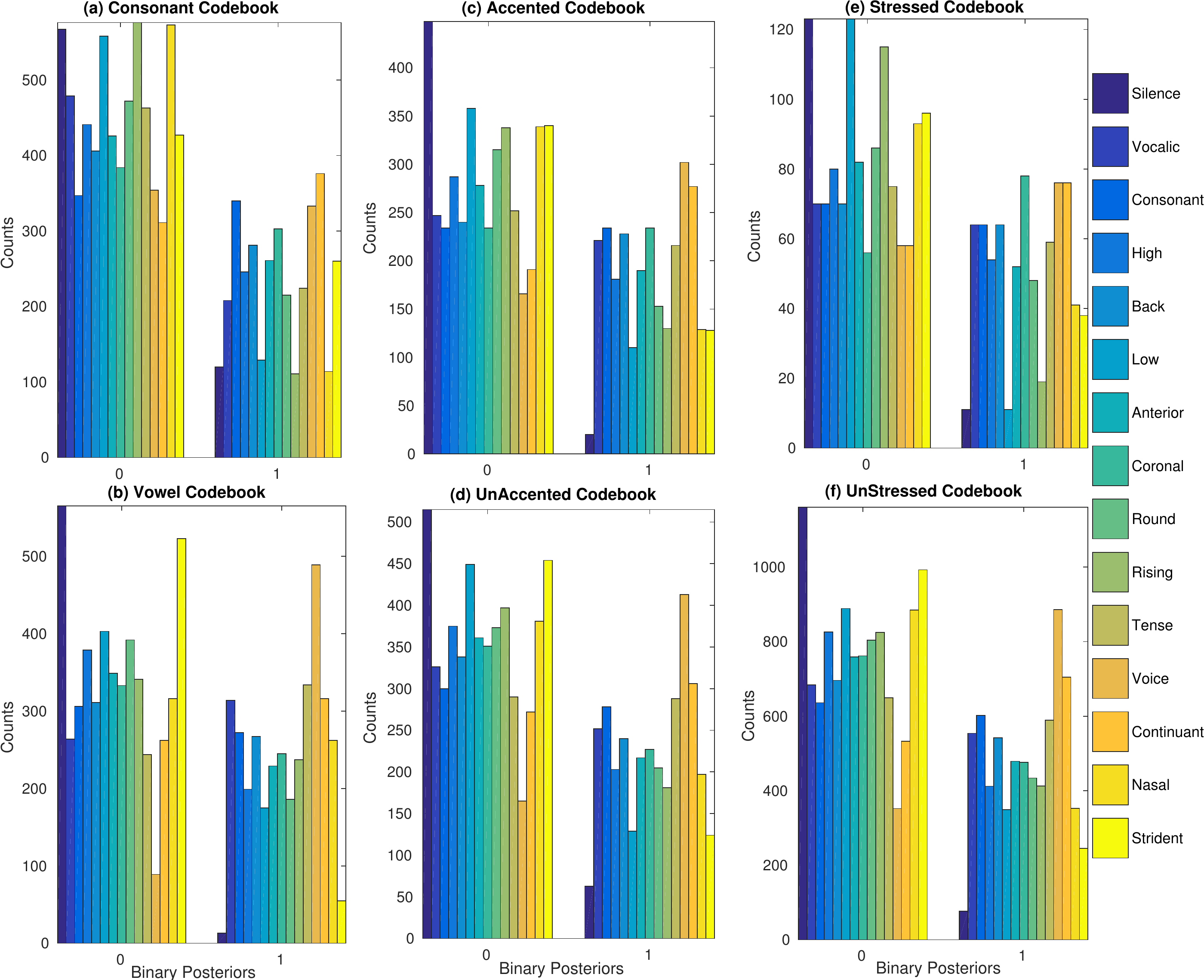}
  \caption{{ \it {\color{mc}Illustration of {\rv English} binary phonological posteriors distributions in each individual codebook. The distributions are calculated for the entire test set of Nancy database. The numbers of codes in Consonant and Vowel codebooks are 687 and 578 respectively, in Accented and Unaccented codebooks are 468 and 578 respectively, and in Stressed and Unstressed codebooks are 134 and 1238 respectively. Different distribution of phonological posteriors entangled with the structural differences underlying the support of binary phonological posteriors enables linguistic parsing through binary pattern matching.}}}
  \label{fig:code-dist}
\end{figure}

{\rv The cardinality of each codebook equals the number of unique class-specific binary structures.} The number of unique structures is indeed a 
small fraction of the whole speech data. For example, the
ratio of unique binary structures for the whole Nancy database (16.6
hours of speech) is about 0.08\% of the total number of phonological
posteriors. Figure~\ref{fig:code-dist} illustrates the distribution of
binary phonological posteriors in the individual codebooks along with
the {\rv number of codes}. Comparing the codebook pairs, we can
identify distinct patterns, for example more frequent consonantal
features in the consonantal codebook, and voice features in the vowel
codebook. {\rv Furthermore, we can see the number of codes in Stressed vs Unstressed codebooks are far less balanced compared to the C-V and Accented/Unaccented codebooks. It might imply that the stressed phonological codes are far more restricted, and phonological pattern matching can potentially be an effective method
for this task. This idea is further investigated in~\cite{Cernak16is} where a highly competitive emphasis detection system is achieved.}


The detection method relies on binary pattern matching and the
codebook with a member which possesses maximum similarity to the
phonological posterior determines its supra-segmental linguistic
property, i.e. being a consonant or vowel, stressed or unstressed and
accented or unaccented. The three parsing scenarios are tested
separately so the linguistic parsing amounts to a binary
classification problem.

We process each speech segment independently. To obtain a decision for
the supra-segmental events from the segmental labels, the labels of
all the segments comprising a supra-segmental event are pulled to form
a decision based on majority counting. In other words, the number of
segments being recognized as a particular event is counted, and the
final supra-segmental label is decided according to the maximum
count. If the similarities of a binary phonological posterior to both
codebooks are equal, the segment is not labeled.

Since we devise a top-down parsing mechanism, we use the
knowledge of supra-segmental boundaries to determine the underlying
linguistic event. {\rv Beyond the supra-segmental boundaries, no other information is available, and the linguistic event is classified merely based on majority count of the segments classified by pattern matching within the a priori known boundary.}

\begin{figure}
\centering
  \includegraphics[width=1\linewidth]{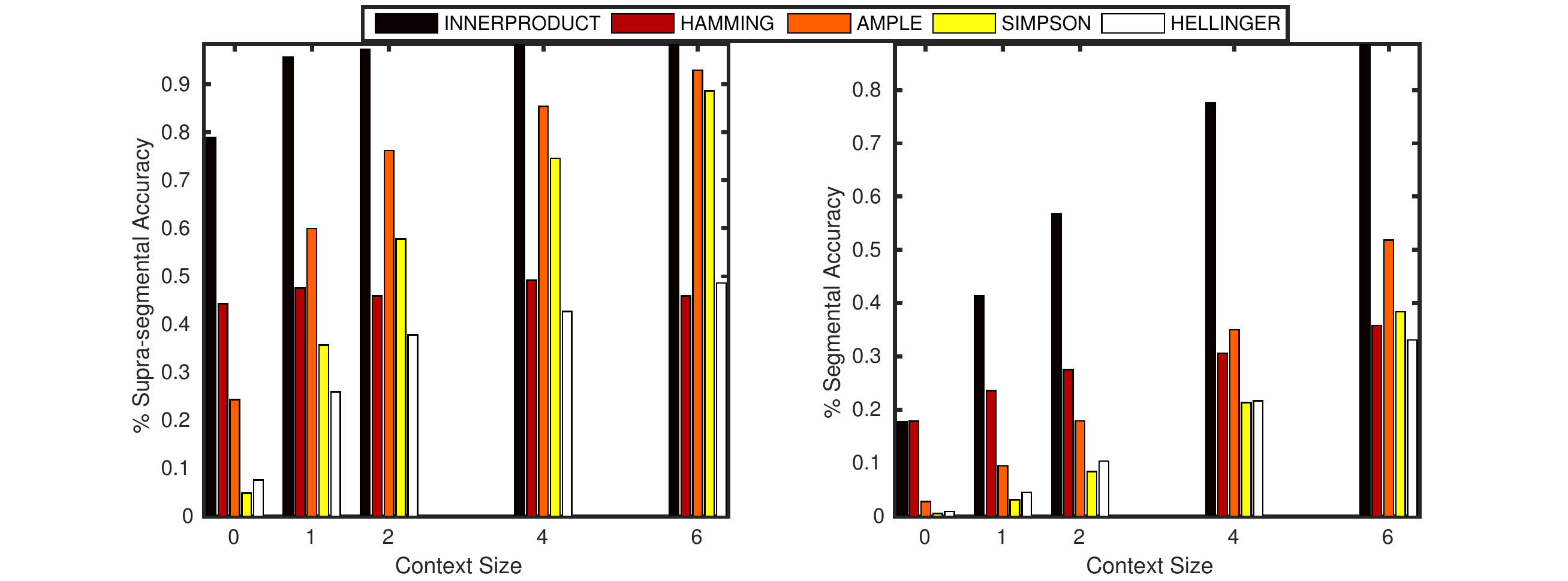}
  \caption{{\it Comparison of the performance of accent detection using various binary similarity measures. The measures are selected from~\citep{Choi10asurvey}. The results of Jaccard~\eqref{eq:jac} is the same as innerproduct. {\color{mc}Nancy database is used for this evaluation. The segmental accuracy corresponds to short segment level accuracy. The segmental decisions are then pooled to make the supra-segmental decisions based on majority counting and exploiting the known boundaries.}}}
  \label{fig:bin-sim}
\end{figure}

To perform pattern matching, the similarity measure of binary
structures must be quantified. There are many metrics formulated for
this purpose~\citep{Choi10asurvey} which differ mainly in the way that
positive/negative match or different mismatches are addressed. We
conducted thorough tests on the metrics defined
in~\citep{Choi10asurvey}; Figure~\ref{fig:bin-sim} compares and
contrasts a few representative results.

We can see that the fast and simple \emph{innerproduct} is the most
effective similarity metric; it quantifies the positive and negative
matches between the two binary structures. On the other hand, Hamming
similarity measure that quantifies the mismatches does not perform
well for linguistic parsing. The Jaccard~\eqref{eq:jac} formula yields
similar results to innerproduct. Hence, we choose the innerproduct for
its efficiency in our linguistic parsing
evaluation. Table~\ref{tab-nancy} lists the accuracy of different
parsing scenarios for English data provided in recordings from Nancy
and French data available in SIWIS database.

{\small
\begin{table}[h]
\centering
\caption{\it Accuracy (\%) of linguistic parsing using structured sparsity pattern matching with different context sizes. The results are evaluated on Nancy and SIWIS speech recordings. The binary similarity measure is innerproduct.}
\label{tab-nancy}
\begin{tabular}{|c|c|c|c|c|c|c|}
\hline
\multicolumn{2}{|c|}{Task \hspace{+0.5mm} / \hspace{+0.5mm} Context Size}        & 0    & 1    & 2    & 4    & 6    \\ \hline \hline
\multirow{2}{*}{C-V Detection}    & Nancy & 53.5 & 83.3 & 88.2 & 93.9 & 96.7 \\ \cline{2-7} 
                                  & SIWIS & 64.5 & 82.9 & 85.5 & 87.9 & 90.3 \\ \hline \hline
\multirow{2}{*}{Stress Detection} & Nancy & 75.4 & 95.4 & 96.9 & 99.5 & 99.5 \\ \cline{2-7} 
                                  & SIWIS & 96.9 & 98.5 & 98.5 & 98.5 & 98.5 \\ \hline \hline
\multirow{2}{*}{Accent Detection} & Nancy & 78.4 & 96.8 & 97.3 & 98.4 & 99.5 \\ \cline{2-7} 
                                  & SIWIS & 91.6 & 93.7 & 93.7 & 94.8 & 94.8 \\ \hline
\end{tabular}
\end{table}
}

The results are averaged over 5-fold random selection of length 1000
consecutive segments. {\color{mc}Accordingly, the codebooks are constructed using the selected data.} The high-order structured sparsity patterns are
obtained by concatenating each segment with its adjacent segments on
the right, and the context size denotes the number of extra segments
concatenated. We can see that the higher order structured sparsity
patterns enables more accurate linguistic parsing. It also confirms
that the proposed structured sparsity principle is independent of
language as well as phonological class definitions.

The differences in stress and accent detection for English and French
might be related to our test data and their ground truth
labels. Although French stress prediction from text is probably
easier, there might {\rv still be some mismatch} between the labels and
recordings. In addition, {\rv the SIWIS} recordings used for
evaluation of prosodic parsing typically contained one emphasized word
per utterance.
Because stress is the relative
prominence within words and utterances, the SIWIS speakers might
emphasize certain words to express word prominence that could lead to
de-emphasis of some syllable prominences we evaluated. The lower
performance of C-V detection might be caused by lower performance of
consonantal- and vocalic-related DNNs, {\rv that is partially
  compensated by longer context. On the other hand, stress and accent
  detection perform better at the segmental level, and relative
  improvement by longer context is smaller than in C-V detection.}
Overall performance of linguistic parsing can be improved by adding
supplementary features, such as features related to tonal and energy
contours.

\subsubsection{Dependency of Linguistic Events}\label{sec:dependency}
Finally, we test  the dependency between different supra-segmental
attributes captured in codebook structures. Both stressed and accented
syllables convey similar information on linguistic emphasis, the
former denotes it at a lexical level while the latter designates it at
a prosodic level. Hence, we hypothesize that the codebook constructed
from stressed structures can be used for accent detection, and vice
versa. Table~\ref{tab:cross} lists the accuracies using these
linguistically relevant codebooks.

{\small
\begin{table}[h]
\centering
\caption{\it Accuracy (\%) of parsing using linguistically relevant codebooks. Namely, we perform stress / accent detection using accent/ stress codebooks to study dependency of stressed and accented structures. } 
\label{tab:cross}
\begin{tabular}{|c|c|c|c|c|c|}
\hline
Task \hspace{+0.5mm} / \hspace{+0.5mm} Context Size & 0    & 1    & 2    & 4    & 6    \\ \hline\hline
Stress detection using accent codebooks       & 63.6 & 82.0 & 79.5 & 82.2 & 85.4 \\ \hline
Accent detection using stress codebooks      & 65.4 & 80.0 & 79.5 & 82.2 & 85.4 \\ \hline

\end{tabular}
\end{table}
}

{\color{ib}We can see that a codebook constructed from either of
  stress/accent structures can be used for detection of the other with
  high accuracy. This study confirms the hypothesis that codebooks
  encapsulate linguistically relevant structures and demonstrates
  that accented structures are indeed highly correlated with the
  stressed structures.}

\section{Concluding Remarks}
\label{sec:concl}
The theories of linguistics and cognitive neuroscience suggest that
the phonological representation of speech places at the heart of
speech temporal organization. We devised a methodology to
quantify the phonological based supra-segmental primitives as
essential building blocks for detection of various linguistic
events. Our proposed approach relies on the identification of
structured sparsity patterns to learn class-specific codebooks
characterizing different supra-segmental attributes. The experiments
confirmed that indeed phonological posteriors convey supra-segmental
information which is encoded in their support of active components,
and these structures can be used as indicators of their higher level
linguistic attributes.

In this context, we also verified that the class-specific structures
of phonological posteriors is a property independent of language as
well as definition of different phonological classes. In addition, it
is robust to unconstrained and noisy recording
conditions. Furthermore, the dependency of different linguistic
properties such as stress and accent is captured in their codebooks
which confirm the high correlation between their underlying
structures. Indeed, the stress and accent TTS labels are related, for
example, for English the placement of accents on stressed syllables in
all content words is a reasonable approximation achieving high
accuracy on typical
databases~\footnote{\url{http://www.festvox.org/bsv/x1750.html}}. In
general, we cannot draw broad conclusions on presented linguistic
parsing, as we {\rv do not have ground truth} reference labels (the
labels predicted from text {\rv do not} necessarily match speech
recordings), and we parsed only a few linguistic classes. However, the
goal of this study has not been to present a complete linguistic
parser, rather we focused on showing supra-segmental properties of
phonological posteriors.

This work quantified the supra-segmental events through the binary
representation of posteriors. This quantification can be more accurate
if multi-level discretization is considered to find a compromise
between speaker and environmental variability encoded in the
probabilities and the actual contribution of phonological classes.

In our future work, we plan to investigate more closely the
relationship of the trajectories of the articulatory-bound
phonological posterior features to the task dynamic model of
inter-articulator coordination in speech~\citep{Saltzman89}. This
study will strengthen our knowledge about the interpretation of
phonological posteriors, when applied to different speech processing
tasks. Applications include detection of syllable boundaries and
subsequent bottom-up linguistic parsing (i.e., parsing without
providing the segment boundaries as discussed by
\cite{Ghitza11,Giraud12}), as well as phonetic posterior estimation
for automatic speech recognition and synthesis systems, parametric
speech coding, and automatic assessment of speech production.

\section{Acknowledgment}
Afsaneh Asaei is supported by funding from SNSF project on ``Parsimonious Hierarchical Automatic Speech Recognition (PHASER)'' grant agreement number 200021-153507. {\color{mc}The authors are grateful to the anonymous reviewers for their time and comments to improve the clarity and quality of the manuscript.}

\section{References}
\label{sec:ref}

\bibliographystyle{elsarticle/elsarticle-harv}
\bibliography{refs}

\begin{thebibliography}{54}
\expandafter\ifx\csname natexlab\endcsname\relax\def\natexlab#1{#1}\fi
\expandafter\ifx\csname url\endcsname\relax
  \def\url#1{\texttt{#1}}\fi
\expandafter\ifx\csname urlprefix\endcsname\relax\def\urlprefix{URL }\fi

\bibitem[{Asaei et~al.(2015)Asaei, Cernak, and Bourlard}]{Asaei15compr}
Asaei, A., Cernak, M., Bourlard, H., Sep. 2015. {On Compressibility of Neural
  Network Phonological Features for Low Bit Rate Speech Coding}. In: Proc. of
  Interspeech. pp. 418--422.

\bibitem[{Bauman-Waengler(2011)}]{Bauman2011}
Bauman-Waengler, J., Mar. 2011. {Articulatory and Phonological Impairments: A
  Clinical Focus (4th Edition) (Allyn \& Bacon Communication Sciences and
  Disorders)}, 4th Edition. Pearson.

\bibitem[{Black et~al.(1997)Black, Taylor, and Caley}]{festival}
Black, A., Taylor, P., Caley, R., 1997. {The Festival Speech Synthesis System}.
  Technical report, Human Communication Research Centre, University of
  Edinburgh.

\bibitem[{Bouchard et~al.(2013)Bouchard, Mesgarani, Johnson, and
  Chang}]{Bouchard13}
Bouchard, K.~E., Mesgarani, N., Johnson, K., Chang, E.~F., Mar. 2013.
  {Functional organization of human sensorimotor cortex for speech
  articulation.} Nature 495~(7441), 327--332.

\bibitem[{Browman and Goldstein(1986)}]{Browman86}
Browman, C.~P., Goldstein, L.~M., May 1986. {Towards an articulatory
  phonology}. Phonology 3, 219--252.

\bibitem[{Browman and Goldstein(1988)}]{Browman88}
Browman, C.~P., Goldstein, L.~M., 1988. {Some Notes on Syllable Structure in
  Articulatory Phonology}. Phonetica 45, 155--180.

\bibitem[{Browman and Goldstein(1989)}]{Browman89}
Browman, C.~P., Goldstein, L.~M., 1989. {Articulatory gestures as phonological
  units}. Phonology 6, 201--251.

\bibitem[{Browman and Goldstein(1992)}]{Browman92}
Browman, C.~P., Goldstein, L.~M., 1992. {Articulatory phonology: An overview}.
  Phonetica 49, 155--180.

\bibitem[{Cernak et~al.(2016{\natexlab{a}})Cernak, Asaei, Honnet, Garner, and
  Bourlard}]{Cernak16is}
Cernak, M., Asaei, A., Honnet, P.-E., Garner, P.~N., Bourlard, H.,
  2016{\natexlab{a}}. {Sound Pattern Matching for Automatic Prosodic Event
  Detection}. In: Proc. of Interspeech. San Francisco, CA, USA.

\bibitem[{Cernak et~al.(2016{\natexlab{b}})Cernak, Benus, and
  Lazaridis}]{Cernak2016a}
Cernak, M., Benus, S., Lazaridis, A., 2016{\natexlab{b}}. Speech vocoding for
  laboratory phonology.
\newline\urlprefix\url{http://arxiv.org/abs/1601.05991}

\bibitem[{Cernak and Garner(2016)}]{Cernak2016phonvoc}
Cernak, M., Garner, P.~N., 2016. {PhonVoc: A Phonetic and Phonological Vocoding
  Toolkit}. In: Proc. of Interspeech. San Francisco, CA, USA.

\bibitem[{Cernak et~al.(2015{\natexlab{a}})Cernak, Garner, Lazaridis, Motlicek,
  and Na}]{Cernak15ieee}
Cernak, M., Garner, P.~N., Lazaridis, A., Motlicek, P., Na, X., Jun.
  2015{\natexlab{a}}. {Incremental Syllable-Context Phonetic Vocoding}.
  {IEEE/ACM Trans. on Audio, Speech, and Language Processing} 23~(6),
  1019--1030.

\bibitem[{Cernak et~al.(2015{\natexlab{b}})Cernak, Potard, and
  Garner}]{Cernak15icassp}
Cernak, M., Potard, B., Garner, P.~N., Apr. 2015{\natexlab{b}}. {Phonological
  vocoding using artificial neural networks}. In: Proc. of {ICASSP}. IEEE, pp.
  4844--4848.

\bibitem[{Choi and Cha(2010)}]{Choi10asurvey}
Choi, S.-s., Cha, S.-h., 2010. A survey of binary similarity and distance
  measures. Journal of Systemics, Cybernetics and Informatics, 43--48.

\bibitem[{Chomsky and Halle(1968)}]{chomsky68sound}
Chomsky, N., Halle, M., 1968. The Sound Pattern of English. Harper \& Row, New
  York, NY.

\bibitem[{Dunn and Everitt(1982)}]{Taxonomy}
Dunn, G., Everitt, B.~S., 1982. {An Introduction to Mathematical Taxonomy}.
  Cambridge University Press.

\bibitem[{Fowler et~al.(2015)Fowler, Shankweiler, and
  Studdert-Kennedy}]{Fowler15}
Fowler, C.~A., Shankweiler, D., Studdert-Kennedy, M., Aug. 2015. {Perception of
  the Speech Code Revisited: Speech Is Alphabetic After All.} Psychological
  review.

\bibitem[{Galliano et~al.(2006)Galliano, Geoffrois, Gravier, f.~Bonastre,
  Mostefa, and Choukri}]{ester06}
Galliano, S., Geoffrois, E., Gravier, G., f.~Bonastre, J., Mostefa, D.,
  Choukri, K., 2006. Corpus description of the ester evaluation campaign for
  the rich transcription of french broadcast news. In: In Proceedings of the
  5th international Conference on Language Resources and Evaluation (LREC 2006.
  pp. 315--320.

\bibitem[{Ghitza(2011)}]{Ghitza11}
Ghitza, O., 2011. {Linking speech perception and neurophysiology: speech
  decoding guided by cascaded oscillators locked to the input rhythm.}
  Frontiers in psychology 2.

\bibitem[{Giraud and Poeppel(2012)}]{Giraud12}
Giraud, A.-L.~L., Poeppel, D., Apr. 2012. {Cortical oscillations and speech
  processing: emerging computational principles and operations.} Nature
  neuroscience 15~(4), 511--517.

\bibitem[{Goldstein and Fowler(2003)}]{Goldstein03}
Goldstein, L., Fowler, C., 2003. {Articulatory phonology: a phonology for
  public language use}. In: Meyer, A., N.Schiller (Eds.), {Phonetics and
  Phonology in Language Comprehension and Production: Differences and
  Similarities}. New York: Mouton, pp. 159--207.

\bibitem[{Harris and Lindsey(1995)}]{Harris95}
Harris, J., Lindsey, G., 1995. {The elements of phonological representation}.
  Longman, Harlow, Essex, pp. 34--79.

\bibitem[{Hickok and Poeppel(2007)}]{Hickok07}
Hickok, G., Poeppel, D., May 2007. {The cortical organization of speech
  processing}. Nature Reviews Neuroscience 8~(5), 393--402.

\bibitem[{Hinton et~al.(2006)Hinton, Osindero, and Teh}]{Hinton06}
Hinton, G.~E., Osindero, S., Teh, Y.~W., Jul. 2006. {A Fast Learning Algorithm
  for Deep Belief Nets}. Neural Comput. 18~(7), 1527--1554.

\bibitem[{Jakobson and Halle(1956)}]{Jakobson56}
Jakobson, R., Halle, M., 1956. {Fundamentals of Language}. The Hague: Mouton.

\bibitem[{Jun(2005)}]{Jun05}
Jun, S.-A., 2005. {Prosodic Typology}. In: Jun, S.-A. (Ed.), {Prosodic
  Typology: The Phonology of Intonation and Phrasing}. Oxford University Press,
  pp. 430--458.

\bibitem[{Ladefoged and Johnson(2014)}]{Ladefoged14}
Ladefoged, P., Johnson, K., Jan. 2014. {A Course in Phonetics}, 7th Edition.
  Cengage Learning.

\bibitem[{Lakatos et~al.(2005)Lakatos, Shah, Knuth, Ulbert, Karmos, and
  Schroeder}]{Lakatos05}
Lakatos, P., Shah, A.~S., Knuth, K.~H., Ulbert, I., Karmos, G., Schroeder,
  C.~E., Sep. 2005. {An oscillatory hierarchy controlling neuronal excitability
  and stimulus processing in the auditory cortex.} Journal of neurophysiology
  94~(3), 1904--1911.

\bibitem[{Lee et~al.(2005)Lee, Yildirim, Kazemzadeh, and Narayanan}]{lee05}
Lee, S., Yildirim, S., Kazemzadeh, A., Narayanan, S., 2005. {An Articulatory
  study of emotional speech production}. In: Proc. of Interspeech. pp.
  497--500.

\bibitem[{Leonard et~al.(2015)Leonard, Bouchard, Tang, and Chang}]{Leonard2015}
Leonard, M.~K., Bouchard, K.~E., Tang, C., Chang, E.~F., 2015. Dynamic encoding
  of speech sequence probability in human temporal cortex. The Journal of
  Neuroscience 35~(18), 7203--7214.

\bibitem[{Leong et~al.(2014)Leong, Stone, Turner, and Goswami}]{Leong14jasa}
Leong, V., Stone, M.~A., Turner, R.~E., Goswami, U., Jul. 2014. {A role for
  amplitude modulation phase relationships in speech rhythm perception.} J.
  Acoust. Soc. Am. 136~(1), 366--381.

\bibitem[{Levelt(1993)}]{Levelt93}
Levelt, W. J.~M., Aug. 1993. {Speaking: From Intention to Articulation (ACL-MIT
  Series in Natural Language Processing)}. A Bradford Book.

\bibitem[{Liberman and Whalen(2000)}]{Liberman00}
Liberman, A.~M., Whalen, D.~H., May 2000. {On the relation of speech to
  language.} Trends in cognitive sciences 4~(5), 187--196.

\bibitem[{Matt(2014)}]{Gordon2014}
Matt, G., 2014. Disentangling stress and pitch accent: Toward a typology of
  prominence at different prosodic levels. in Harry van der Hulst (ed.). To
  appear, In Word Stress: Theoretical and Typological Issues, Oxford University
  Press.

\bibitem[{Mesgarani et~al.(2014)Mesgarani, Cheung, Johnson, and
  Chang}]{Mesgarani2014}
Mesgarani, N., Cheung, C., Johnson, K., Chang, E.~F., Feb. 2014. {Phonetic
  Feature Encoding in Human Superior Temporal Gyrus}. Science 343~(6174),
  1006--1010.

\bibitem[{Miller and Nicely(1955)}]{Miller55}
Miller, G.~A., Nicely, P.~E., Mar. 1955. {An Analysis of Perceptual Confusions
  Among Some English Consonants}. J. Acoust. Soc. Am. 27~(2), 338--352.

\bibitem[{Nagamine et~al.(2015)Nagamine, Seltzer, and Mesgarani}]{Nagamine15}
Nagamine, T., Seltzer, M.~L., Mesgarani, N., 2015. Exploring how deep neural
  networks form phonemic categories. In: Proc. of Interspeech. pp. 1912--1916.

\bibitem[{Nam et~al.(2009)Nam, Goldstein, and Saltzman}]{Nam09}
Nam, H., Goldstein, L., Saltzman, E., 2009. {Self-organization of syllable
  structure: A coupled oscillator model}. In: Pellegrino, F., Marisco, E.,
  Chitoran, I. (Eds.), {Approaches to phonological complexity}. Berlin, New
  York: Mouton de Gruyter, pp. 299--328.

\bibitem[{Paul and Baker(1992)}]{WSJDB}
Paul, D.~B., Baker, J.~M., 1992. {The design for the wall street journal-based
  CSR corpus}. In: Proceedings of the workshop on Speech and Natural Language.
  HLT '91. Association for Computational Linguistics, Stroudsburg, PA, USA, pp.
  357--362.

\bibitem[{Perennou(1986)}]{bdlex}
Perennou, G., 1986. {B.D.L.E.X.} : A data and cognition base of spoken
  {F}rench. In: Proc. of {ICASSP}. Vol.~11. pp. 325--328.

\bibitem[{Phillips et~al.(2000)Phillips, Pellathy, Marantz, Yellin, Wexler,
  Poeppel, McGinnis, and Roberts}]{Phillips00}
Phillips, C., Pellathy, T., Marantz, A., Yellin, E., Wexler, K., Poeppel, D.,
  McGinnis, M., Roberts, T., Nov. 2000. {Auditory Cortex Accesses Phonological
  Categories: An MEG Mismatch Study}. Journal of Cognitive Neuroscience 12~(6),
  1038--1055.

\bibitem[{Poeppel(2003)}]{Poeppel03}
Poeppel, D., Aug. 2003. {The Analysis of Speech in Different Temporal
  Integration Windows: Cerebral Lateralization As 'Asymmetric Sampling in
  Time'}. Speech Communication 41~(1), 245--255.

\bibitem[{Poeppel(2014)}]{Poeppel14}
Poeppel, D., Oct. 2014. {The neuroanatomic and neurophysiological
  infrastructure for speech and language}. Current Opinion in Neurobiology 28,
  142--149.

\bibitem[{Povey et~al.(2011)Povey, Ghoshal, Boulianne, Burget, Glembek, Goel,
  Hannemann, Motlicek, Qian, Schwarz, Silovsky, Stemmer, and
  Vesely}]{PoveyASRU2011}
Povey, D., Ghoshal, A., Boulianne, G., Burget, L., Glembek, O., Goel, N.,
  Hannemann, M., Motlicek, P., Qian, Y., Schwarz, P., Silovsky, J., Stemmer,
  G., Vesely, K., Dec. 2011. The kaldi speech recognition toolkit. In: Proc. of
  ASRU. IEEE SPS, iEEE Catalog No.: CFP11SRW-USB.

\bibitem[{Rasipuram and Magimai.-Doss(2011)}]{Rasipuram11}
Rasipuram, R., Magimai.-Doss, M., May 2011. {Integrating articulatory features
  using Kullback-Leibler divergence based acoustic model for phoneme
  recognition}. In: Proc. of {ICASSP}. IEEE, pp. 5192--5195.

\bibitem[{Roekhaut et~al.(2014)Roekhaut, Brognaux, Beaufort, and
  Dutoit}]{Roekhaut2014}
Roekhaut, S., Brognaux, S., Beaufort, R., Dutoit, T., 2014. e{L}ite-{HTS}: A
  {NLP} tool for {F}rench {HMM}-based speech synthesis. In: Proc. of
  Interspeech. pp. 2136--2137.

\bibitem[{Saltzman and Munhall(1989)}]{Saltzman89}
Saltzman, E.~L., Munhall, K.~G., 1989. {A dynamical approach to gestural
  patterning in speech production}. Ecological Psychology 1, 333--382.

\bibitem[{Shinoda and Watanabe(1997)}]{shinoda:mdl}
Shinoda, K., Watanabe, T., 1997. Acoustic modeling based on the {MDL} principle
  for speech recognition. In: Proc. of Eurospeech. pp. I --99--102.

\bibitem[{Siniscalchi et~al.(2012)Siniscalchi, Lyu, Svendsen, and
  Lee}]{Siniscalchi2012}
Siniscalchi, S.~M., Lyu, D.-C., Svendsen, T., Lee, C.-H., Mar. 2012.
  {Experiments on Cross-Language Attribute Detection and Phone Recognition With
  Minimal Target-Specific Training Data}. IEEE Trans. on Audio, Speech, and
  Language Processing 20~(3), 875--887.

\bibitem[{Stevens(2005)}]{Stevens08}
Stevens, K.~N., 2005. {Features in Speech Perception and Lexical Access}. In:
  Pisoni, D.~B., Remez, R.~E. (Eds.), {The Handbook of Speech Perception}.
  Blackwell Publishing, pp. 125--155.

\bibitem[{Stouten and Martens(2006)}]{Stouten06}
Stouten, F., Martens, J.-P., May 2006. {On The Use of Phonological Features for
  Pronunciation Scoring}. In: Proc. of {ICASSP}. Vol.~1. IEEE, p.~I.

\bibitem[{Wernicke(1874/1969)}]{Wernicke1874}
Wernicke, C., 1874/1969. {Bonston studies in the phylosophy of science}. In:
  Cohen, R., Wartofsky, M. (Eds.), {The symptom complex of aphasia: A
  psychological study on an anatomical basis}. D. Reichel, Dordrecht, pp.
  34--97.

\bibitem[{Yu et~al.(2012)Yu, Siniscalchi, Deng, and Lee}]{Yu2012}
Yu, D., Siniscalchi, S., Deng, L., Lee, C.-H., March 2012. Boosting attribute
  and phone estimation accuracies with deep neural networks for detection-based
  speech recognition. In: Proc. of {ICASSP}. IEEE SPS.

\bibitem[{Zen et~al.(2007)Zen, Nose, Yamagishi, Sako, Masuko, Black, and
  Tokuda}]{Zen:HTS}
Zen, H., Nose, T., Yamagishi, J., Sako, S., Masuko, T., Black, A., Tokuda, K.,
  2007. The {HMM}-based {S}peech {S}ynthesis {S}ystem {V}ersion 2.0. In: Proc.
  of ISCA SSW6. pp. 131--136.

\end{thebibliography}



\end{document}